\documentclass[conference]{IEEEtran}
\usepackage{url}
\IEEEoverridecommandlockouts

\usepackage{cite}
\usepackage{amsmath,amssymb,amsfonts}
\usepackage{algorithmic}
\usepackage{graphicx}
\usepackage{textcomp}
\usepackage{xcolor}
\usepackage{textcomp}
\usepackage{amsmath}
\usepackage{soul}
\usepackage{color}
\usepackage{lipsum}
\usepackage{booktabs}
\usepackage{multirow}
\usepackage{makecell}
\usepackage{graphicx}

\def\BibTeX{{\rm B\kern-.05em{\sc i\kern-.025em b}\kern-.08em
    T\kern-.1667em\lower.7ex\hbox{E}\kern-.125emX}}
\begin{document}

\title{CEAA: A Cognitive Embodied Agents Architecture for Interactive Computing Systems\\}

\author{\IEEEauthorblockN{Aimilios Hadjiliasi} 
\IEEEauthorblockA{\textit{University of Central Lancashire, Cyprus} \\ 
\textit{InSPIRE Research Center} \\ 
AHadjiliasi1@uclan.ac.uk \\
} 

\and
\IEEEauthorblockN{Louis Nisiotis} 
\IEEEauthorblockA{\textit{University of Central Lancashire, Cyprus} \\ 
\textit{InSPIRE Research Center} \\ 
LNisiotis@uclan.ac.uk \\} 

}
\maketitle

\begin{abstract}
The development of embodied Intelligent Virtual Agents (IVAs) that have cognitive capabilities in real-time interactive virtual environments remains a challenge, even with today's advancements in technology. Existing architectures are often focused on either the implementation of low-level reactive control systems that are constrained by commercial game engines, or high-level representations of reasoning models that can be difficult to implement in virtual worlds. This paper builds on that notion and proposes a modular cognitive architecture for deploying embodied IVAs. This architecture builds on existing, pre-established frameworks such as the Sense-Think-Act paradigm and the Belief-Desire-Intention cognitive model, among others, and aims to provide a reusable implementation-oriented framework as a template for deploying IVA “brains” in interactive 3D computing systems. The proposed architecture contributes by providing a modular, implementation-oriented framework for the deployment of embodied, cognitive-capable IVAs and bridges the gap between high-level agent reasoning models with real-time embodied execution, for scalable, adaptive, and explainable agents in complex interactive virtual environments.
\end{abstract}


\section{Introduction}
With the advancement of technology, Intelligent Virtual Agents (IVAs) have become a key component of complex computing systems, ranging from Virtual Reality (VR) environments to Virtual Museums (e.g., \cite{nisiotis2023interwoven}), serious games (e.g., \cite{magnenat2009virtual}), and a plethora of Metaverse applications (e.g., \cite{schmidt2024frankenstein, nisiotis2025developing}). 
This is because agents can guide users and provide adaptive and personalized experiences, among many other functions. 
For example, agents can be used in an educational setting \cite{hadjiliasiNisiotis2026} to guide learners through learning materials and support their learning; they can be used as NPCs in video games for entertainment purposes \cite{takahashi2018adaptive}; and they can generally be used in applications for multiple reasons, including being used as an audience for practicing presentation skills and overcoming public speaking anxiety \cite{takac2019public}, among many others.

Due to the interdisciplinary use of the term agent, this paper is focused on agents that are situated in virtual environments and have embodied representation. 
Depending on the needs of the application, these agents range in terms of capabilities. 
However, in complex systems, especially in systems where agents have more responsibilities rather than simply presenting information, they are expected to operate autonomously, interact naturally with users, and adapt their behavior over time \cite{yang2025embodied, kota2025evolution}. 
To achieve this functionality, the deployment of agents requires architectures that are cognitively expressive and practically deployable in real-time \cite{kishor2025agentic}.
Despite extensive research on intelligent agents and recent advances in AI, there remains a persistent lack of implementation-oriented cognitive architectures capable of unifying high-level reasoning, memory, planning, and real-time embodied action within dynamic interactive environments.

From a historical point of view, many IVAs in virtual environments are implemented on reactive and/or semi-reactive approaches \cite{kota2025evolution, Plch2011TowardsBI}. 
These approaches include the use of state machines, behavior trees, or scripted rule-based systems, due to their efficiency and ease of development in game engines, such as Unity and Unreal Engine. 
These approaches are very effective at controlling moment-to-moment behavior. 
However, they often struggle to adapt to dynamic environments and support goal-directed reasoning, explainability, and personalization \cite{joshi2025evolution}.
Simultaneously, several cognitive architectures that simulate human-like practical reasoning have been proposed, such as the Belief-Desire-Intention (BDI) model \cite{rao1995bdi}, the State-Operator-and-Results (SOAR), and Adaptive Control of Thought—Rational (ACT-R) models, among many others \cite{chong2007integrated}. 
Those provide a rich theoretical foundation for deploying software agents that are capable of reasoning and decision-making. 
However, these architectures tend to be difficult to integrate into real-time 3D environments due to their complexity. 
As a result, the practical feasibility of deploying IVAs with embodied representations in complex virtual environments remained a challenge.

In response to this limitation and as part of wider research investigating the use of pedagogical agents in VR  \cite{hadjiliasi2024comparative}, the Cognitive Embodied Agent Architecture (CEAA) is devised for deploying cognitively capable agents as the backbone for deploying the “brain” of multiple IVAs. 
The proposed architecture blends and builds upon several architectures for deploying agents from different levels of abstraction.  
Specifically, the proposed architecture has as a core component the classical Sense-Think-Act paradigm and extends it by integrating blackboard-based shared knowledge, explicit memory, and memory processor components, and the BDI reasoning model. 
Effectively, this architecture introduces a clear separation between the agent’s cognitive states, reasoning, planning, and embodied actions, and as a subsequent event, it facilitates integration with modern game engines such as Unity and Unreal. 

\section{Background and Context}
\subsection{Intelligent Virtual Agents and Embodiment in Virtual Worlds}
IVAs have been extensively investigated over the past two decades as autonomous or semi-autonomous entities in virtual environments. 
These entities are advanced AI-powered software designed to conduct natural, conventional, and personalized interactions with users in digital systems, and they are capable of perceiving, reasoning, and acting in the virtual world in which they are situated \cite{hadjiliasiNisiotis2026}. 
Early research defined IVAs as software agents with autonomy, reactivity, proactivity, and social capabilities within the human-computer interaction domain \cite{wooldridge1995intelligent}. 
With the advancement of virtual environments and specifically 2D/3D worlds, IVAs have adopted an embodied representation, and they have been used across multiple domains, including entertainment, education, training, and healthcare, among many others \cite{kruse2023would,kyrlitsias2022social,rickel2001intelligent}. 
This enabled them to evolve from abstract decision-making entities to visually existent characters that are situated in multidimensional virtual worlds \cite{cassell2000designing}. 
This embodiment has been investigated over the years, and research demonstrated its ability to significantly affect user engagement, trust, social presence, and perceived intelligence, especially in immersive virtual environments.

However, embodiment comes with some drawbacks. 
In particular, in immersive and interactive virtual environments, embodiment introduces additional requirements and expectations beyond the classical agent intelligence, such as decision-making, suggestions, and assistance, among others \cite{rickel2002toward, cassell2000designing}. 
Embodiment means that the agent has a visible presence in the environment, in any form, but typically it comes in a human-like representation. 
Hence, from the computer graphics perspective, embodied agents must coordinate their cognition with perception, behavior, animation, and communication in real time \cite{nowak2003effect}. 
As a result, this introduces a tight relationship between the agent’s cognitive processes and physical realizations. 
This process has a subsequent event, complicating the development of intelligent behavior, as agents must simultaneously reason about goals and constraints and also respond to continuous user input and environmental dynamics.

\subsection{Agent Architectures and Cognitive Models}
\begin{figure*}[t]
    \centering
    \includegraphics[width=\linewidth]{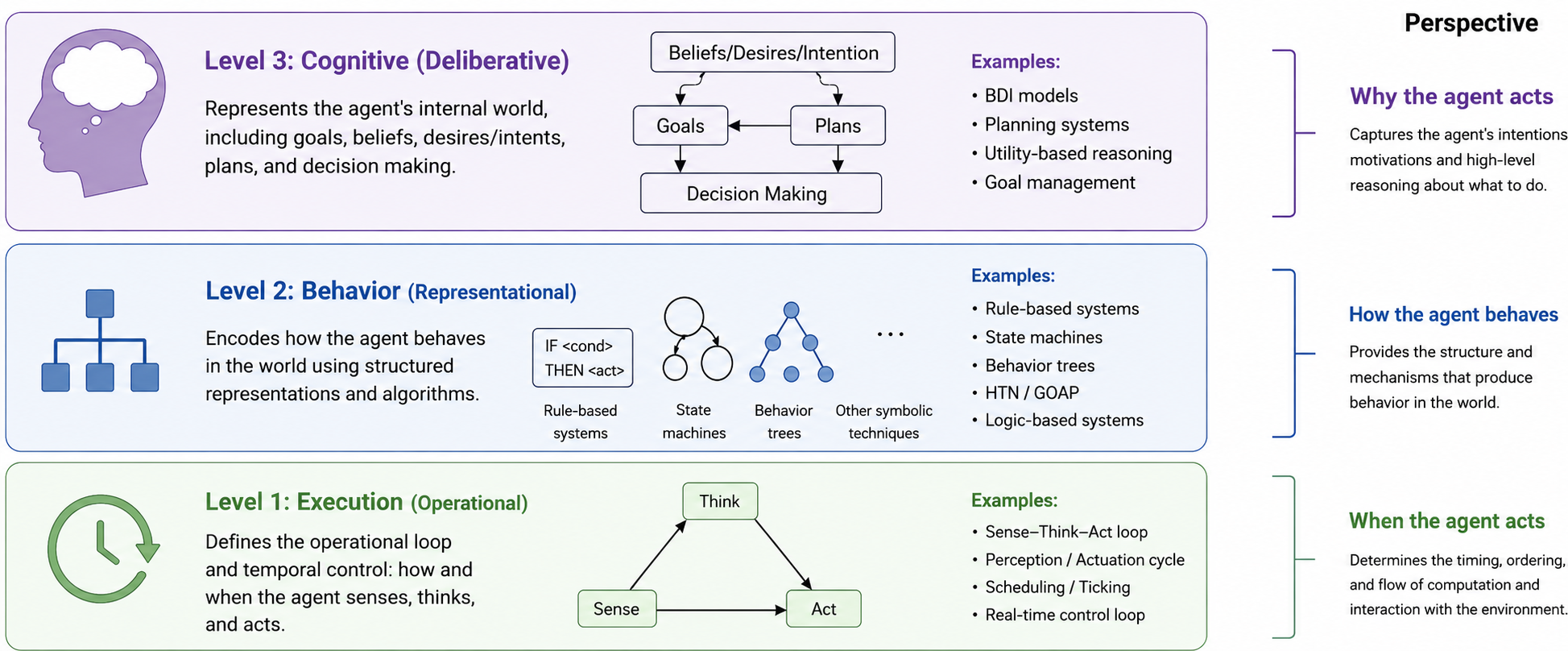}
    \caption{Graphical representation of the three levels of abstraction in agent architectures.}
    \label{fig:agent-architecture-levels}
\end{figure*}

To introduce intelligence and implement agents in virtual environments, many authors proposed several architectures, models, and paradigms over the years. 
Agent architectures are the structural foundation for implementing intelligent behavior in agents \cite{RussellNorvig2020,wooldridge1995intelligent}.
As illustrated in Figure~\ref{fig:agent-architecture-levels}, agent architectures can be analyzed across execution, behavioral and cognitive levels of abstraction.
From a high-level perspective, agent architectures can be viewed as operating across three main levels of abstraction \cite{DeRidder_LayeredAgentArchitectures_2025, ferguson1992touringmachines}. 
These levels explain different aspects of agent deployment, and specifically how agents run, how they behave, and why they act \cite{wooldridge2009introduction, RussellNorvig2020}. 
The lowest level describes the execution and temporal control of the agent, which effectively defines the operational and continuously updated loop of the agent. 
These types of architectures are responsible for defining when computation occurs instead of how decisions are made \cite{yao2022react, Rufus2025AgentLoop, ferguson1992touringmachines}. 
The middle level of abstraction relates to the behavior representations and encompasses mechanisms such as rule-based systems, state machines, behavior trees, and other symbolic AI techniques \cite{AllgeuerBehnke2018, wooldridge2009introduction, iovino2022survey}. 
This layer provides a concrete structure for organizing the behavior of the agent, but typically, it lacks representations of the goals, beliefs, and cognition of the agent. 
The highest level of abstraction focuses on the cognitive capabilities of the agents. 
This level of abstraction explains the why behind the behavior of an agent over time, instead of selecting its individual actions \cite{rao1995bdi, de2020bdi, Khodaygani2025CognitiveModelingAgents}. 
In practice, an effective IVA should be implemented based on all three levels of abstraction. 
In particular, the agent should use the low-level for execution loops to be able to operate in real-time, use the middle-level for structuring its behavior to coordinate actions, and leverage the highest-level of abstraction for cognitive reasoning to provide purpose and adaptability.

One of the fundamental paradigms that runs on both the lowest and middle levels of abstraction is the Sense-Think-Act paradigm. This paradigm conceptualizes agents as systems that perceive their environment, think internally, and execute actions accordingly \cite{chetty2019embodied, siegel2003sense}. 
This paradigm offers conceptual clarity when it comes to the agent’s internal processes.
However, it can sometimes lack sufficient details on how to implement complex reasoning and memory mechanics that are required by agents in dynamic virtual environments.

On the other hand, one of the most popular architectures that has been proposed for the higher level of abstraction is the BDI architecture. 
This architecture was inspired by Michael Bratman’s philosophical model of human practical reasoning and serves as a foundational framework for deploying autonomous agents in intelligent virtual environments. 
The three main components of this architecture are the beliefs, desires, and intentions of the agent. 
The belief component encapsulates the agent’s situated subjective knowledge and perceptions of the world. 
This component is typically represented as a dynamic belief base that can be updated through sensory input. 
The desire component encapsulates the agent’s motivational states or goals it seeks to achieve. 
Last, the intentions component represents the main objectives of the agent, and its cognitive commitment to specific plans or courses of action selected from desires. 
Effectively, this architecture facilitates rational decision-making by integrating perception, option generation, commitment, and plan execution in a cyclical process, making it particularly effective for real-world applications such as robotics, virtual assistants, and simulations where agents must balance reactivity and proactivity \cite{ortiz2022modularization, Perez2019BDI, georgeff1998belief}.

Similar to BDI, several other architectures have been proposed in the scientific literature, including SOAR \cite{laird1987soar} and ACT-R \cite{taatgen2006modeling}. 
These architectures focus on modeling cognition through production rules, memory subsystems, and learning mechanisms, mainly for simulating human intelligence, problem-solving, and cognition \cite{laird1987soar, laird2019soar}. 
Specifically, SOAR operates through a unified theory of cognition based on the problem-space computational model. 
In this model, agents perceive their environments and generate and select their operators based on production rules, something that enables them to perceive, plan, and learn. 
On the other hand, ACT-R focuses on a modular psychological structure of declarative and procedural memory, which enables agents to perceive, control, and manage their goals \cite{laird2022analysis, ritter2019act}. 

\subsection{Knowledge Representation and Memory in Virtual Agents}
A key component of any type of agent is memory and knowledge representation. 
These components determine how they encode, store, and exploit information over time. 
In classical cognitive agent architectures, knowledge is typically represented symbolically. 
Symbolic representation includes facts and rules that are stored in distinct memory sub-systems. 
Some types of memory include episodic and procedural memories, which store the short- and long-term facts needed to be known by the agent \cite{anderson1997act,newell1994unified, RussellNorvig2020}.
Building on this idea, agents can leverage it, and they were provided with capabilities that enable them to learn, personalize based on user behavior, and simulate human memory, which allows them to develop knowledge. 
Even though this represents a great advancement for agents, this capability is hard to implement in real-time, due to performance and complexity constraints \cite{de2011neural, yao2022react, zhang2025survey}. As a result, this highlights the difficulties in practically developing and deploying agents with such capabilities.

\subsection{Embodied IVAs in Commercial Game Engines}
The deployment of IVAs is heavily constrained by the capabilities of modern engines that are used for development, such as Unity and Unreal Engine. 
These engines provide a tool for game development and, to an extent, virtual worlds and agents. 
These engines focus on component-based, object-oriented, and event-driven execution. 
This creates a tight integration between the logic of the environment, the logic of the agent, the physics of the game/simulation, animation, rendering, etc. 
However, even though these engines provide a solid tool for the development of games, simulations, virtual environments, and, of course, agents including their characteristics (embodiment, navigation, behavior, and interaction), at their current state, they offer limited support for the development of high-level cognition, beyond a reactive approach \cite{nystrom2014game,gregory2018game}. 
As a result, existing implementations of IVAs often lack the logic related to cognitive logic within engine-specific components, leading to tightly coupled systems that are difficult to extend, reuse, or reason about.

\section{Proposed System Architecture}
Building on this information, to the best of our knowledge, while there is a significant amount of work on agent architectures as a means to provide cognitive capabilities to agents, this work only provides the conceptual foundation for cognition. 
Cognitive architectures offer rich models of reasoning and memory but lack implementation guidance and real-time feasibility, whereas game AI approaches prioritize performance and embodiment at the expense of cognitive depth and explainability. 
On top of that, current solutions are constrained by the development engine capabilities, and this gap motivated the design and development of CEAA. 

CEAA draws upon agent models originating from different levels of abstraction and is intended to function as a backbone or reusable template for implementing the “brain” of individual embodied agents in virtual environments. 
In particular, the architecture extends the classical sense–think–act paradigm by integrating a cognitive reasoning model based on the BDI framework. 
At the same time, it introduces additional components to explicitly support shared knowledge representation, memory management, planning, and the transformation of abstract decisions into executable behaviors. 
In particular, the architecture is separated into three layers and twelve interconnected components that collectively support perception, reasoning, and action. 

Considering the three layers, the first layer is the \textbf{\textit{User and Environment Layer}} where the users, the agents, and the different virtual materials (depending on the scenario) are situated. 
The second layer is the \textbf{\textit{Knowledge Layer}}. 
This layer is responsible for capturing and maintaining an up-to-date log and state of the environment. 
Specifically, this layer keeps track of all the events happened in the environment and stores them in a meaningful structure so the actions of the users can be seen as a problem that the different agents have to solve. 
The last layer is the \textbf{\textit{Agent Layer}}. 
This layer is responsible for enabling the virtual agent to perform human-like reasoning and decision-making to act in the virtual environment. 

\begin{figure}[h]
    \centering
    \includegraphics[width=\linewidth]{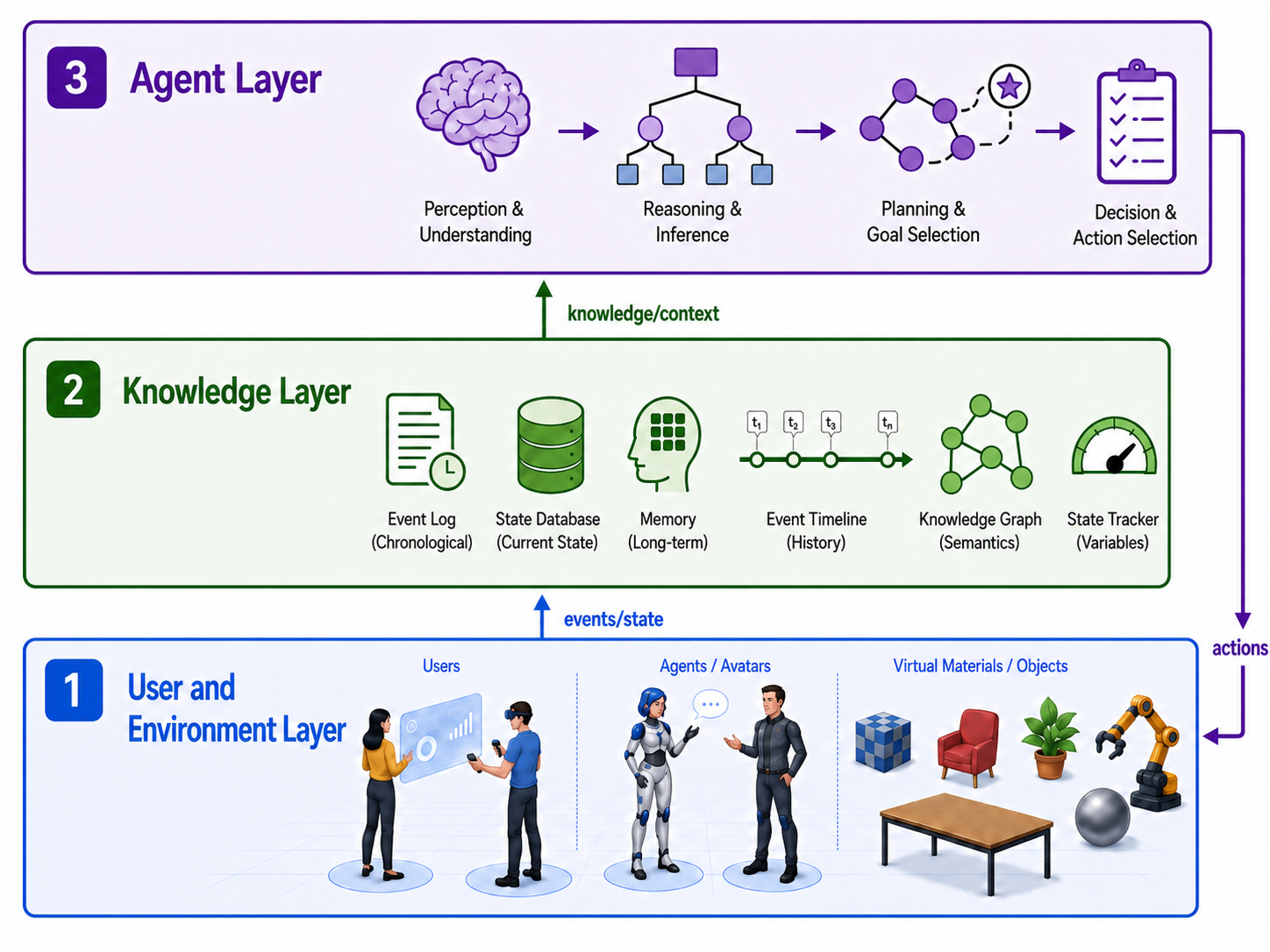}
    \caption{Three-layer virtual agent architecture integrating the environment, knowledge representation, and agent reasoning and action.}
    \label{fig:layers}
\end{figure}

\begin{figure*}[t]
  \centering
  \includegraphics[width=\textwidth]{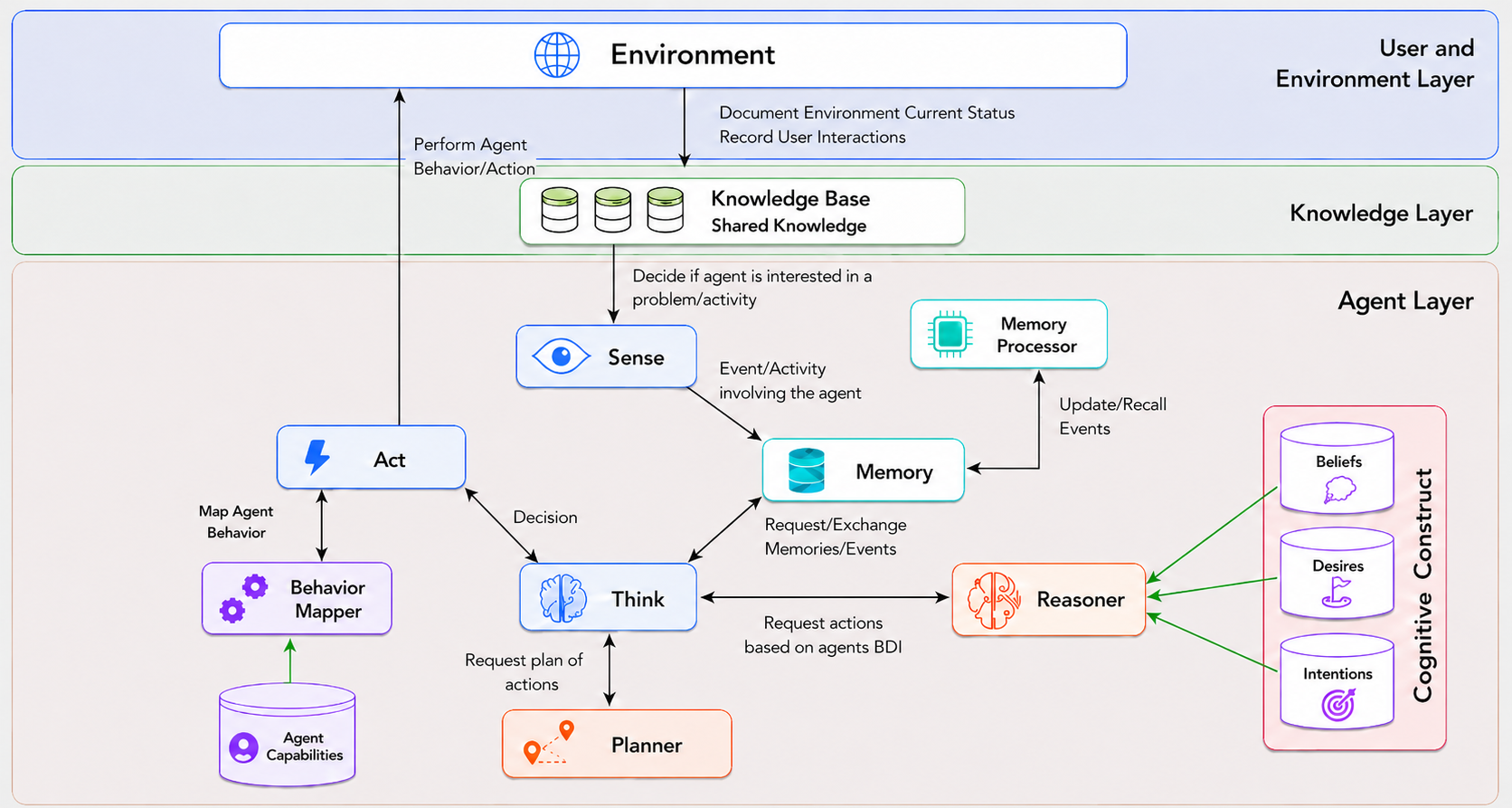}
  \caption{High-level overview of CEAA architecture.}
  \label{fig:CEAA}
\end{figure*}

Considering the components of this architecture, they are the \textit{Environment} interface, a \textit{Knowledge Base} implemented as a shared blackboard between the different agents of that are situated in the environment, a \textit{Sense} component for selective perception, a \textit{Memory} component and associated \textit{Memory Processor} for experience management, a \textit{Think} component for cognitive orchestration, a \textit{Cognitive Construct} reasoner operating over \textit{Belief, Desire, and Intention} structures, a \textit{Reasoner} to perform informed decisions based of the Cognitive Constructs of the agent, a \textit{Planner} for generating action sequences, an \textit{Agent Capabilities} model defining executable affordances, a \textit{Behavior Mapper} for grounding abstract actions, and an \textit{Act} component responsible for executing behaviors within the virtual environment. 
A high-level overview of the proposed architecture is provided in Figure ~\ref{fig:CEAA}.

\textbf{\textit{Environment Component}}: represents the external virtual world in which users exist, and agents are situated. 
It includes all dynamic elements with which users may interact, including the users themselves, other agents, virtual objects with digital meaning, and system-level events relevant only to internal system processes.
This component is not cognitive and performs no reasoning or decision-making. 
Instead, it provides stimuli to the agents. 
Changes caused by user interactions, events of interest, other agents, or system processes generate potentially relevant events. 
These are not interpreted directly by the agent but are recorded in the architecture’s second component, the Shared Knowledge Base.
Events may be captured through different techniques depending on the environment and available hardware. 
These include event-driven publish–subscribe mechanisms, simple if–else statements, and symbolic AI methods such as grid-based planning for detecting events within specific environmental cells. 
Such mechanisms enable real-time detection of user actions, object-state changes, agent movement, collisions, and other system-level events. 
Once detected, each event is converted into structured information and passed to the Shared Knowledge Base.

\textbf{\textit{Knowledge Base Component}}:
This component maintains a structured record of environmental events and the current state of the virtual world. 
It functions as a centralized shared knowledge structure based on the blackboard architecture paradigm, a symbolic AI approach in which multiple independent and specialized knowledge sources collaboratively solve complex problems. 
The paradigm can be compared to several people working around a shared blackboard, with each contributing only to the parts they understand or consider relevant.
Collectively, their contributions support problem resolution \cite{carver1994evolution}.
Following this principle, the Knowledge Base records environmental events and states as shared problems that agents can selectively address according to their goals through their Sense components. 
It therefore maintains an up-to-date symbolic representation of the environment and a structured log of recent events. 
The component also acts as an intermediary between the environment and the agents’ internal processes, allowing multiple cognitive components to access and contribute information without direct dependencies. 

\textbf{\textit{Sense Component}}:
This component is responsible for enabling the agent to sense information through sensors. 
These sensors act on the Knowledge Base layer and specifically in the blackboard model. 
These sensors “sense” a problem that the agent is interested in. 
If a problem exists on the blackboard (e.g., the user requested a presentation by an agent in the virtual environment), the agent, if it is interested, attempts to solve it. 
To solve the problem, the relevant information about the problem is passed to the Memory component of the agent to examine whether similar or relevant things happened in the environment, which only involved this particular agent. 
In essence, this component represents the perceptual mechanisms of the agent, and its responsibility is to determine which events or environment state changes are relevant to the agent. 
This selective attention enables the agent to prevent overloads and ensure that higher-level reasoning processes are triggered after meaningful stimuli. 
Additionally, this component evaluates events based on specific criteria of interest, such as roles, beliefs, desires, and intentions, and makes the agent focus only on the processes of interest. 
To assess whether information is relevant to the agent, simple rule-based filters can be used, along with pandemonium-inspired algorithms. 
Subsequently, once an event is deemed relevant to the agent, it is forwarded to the Memory component for storage.

\textbf{\textit{Memory Component}}:
This component is responsible for storing and organizing the agent’s past experiences in a structured way, so the agent can effectively acquire knowledge. 
This component provides the required continuity that is required for the agent to adapt in the environment based on user behavior and effectively perform context-aware behavior. 
On top of the episodic information, this component also encompasses the semantic knowledge of the agent, including facts, concepts, and user models that are required to shape the personality of the agent. 
Furthermore, this component also encompasses the current understanding of the agent of the world, by constantly communicating with the Think and Memory Processor components. 
With the Memory Processor component, this component shares the episodic events that happened with the agent and retrieves actions that have been previously performed by the agent. 
As a result, knowledge can be created and stored in this component. 
On the other hand, with the Think component, this component responds to a request for past actions. 
Last, the algorithms that are associated with this component focus on storage, indexing, and retrieval, including similarity-based recall, context matching, and case-based reasoning (e.g., by using state machines). 

\textbf{\textit{Think Component}}:
This component serves as the central cognitive coordinator and as an orchestration component of the agent. 
This component is responsible for integrating the information from the Memory component, and based on that information, the agent communicates with the other modules to decide which is the appropriate action to take. 
In the same way, this component also makes requests to the Memory component to examine whether similar events have occurred in the environment. 
As an orchestrator, this component has bidirectional communication with the Reasoner and the Planner components to examine whether an action is aligned with the cognitive processes of the agent and plan the most appropriate actions/outputs. 
This communication enables the agent to express deep reasoning and decision-making. 
However, this component does not implement a specific reasoning algorithm by itself. 
Instead, it orchestrates the flow of information between the cognitive modules and the past experiences of the agent. 
In a nutshell, when an event triggers the agent to act, this component retrieves the beliefs, desires, and intentions of the agent, along with its current knowledge and memories, and proceeds to a decision-making and reasoning process to evaluate the agent’s goals. 
Subsequently, when a decision is made, this component communicates with the planner component to plan which is the most appropriate action to take. 
As a result, this component embodies the control logic of the agent to ensure that perception, memory, reasoning, and planning are combined in a consistent manner.

\textbf{\textit{Cognitive Construct Component}}:
This component comprises three database-like sub-components. 
Those are the agent's Beliefs, Desires, and Intentions. 
Those represent the agent’s core internal mental state and shape its behavior. 
Although conceptually distinct, they provide continuity across reasoning cycles and collectively form a stable, adaptive, and explainable framework for goal-oriented behavior. 
Each is initially programmed to establish the agent’s starting state but can be dynamically updated at run-time as the agent interacts with its environment. 
Search and decision-making algorithms may be used to manage these structures.
\textit{Beliefs} represent the agent’s current understanding of the world, including assumptions about environmental states, user knowledge and behavior, and task processes.
They may be symbolic, probabilistic, or weighted by confidence values, enabling the agent to reason under uncertainty and revise its understanding when new information becomes available.
On the other hand, \textit{Desires} represent the agent’s goals and motivations, which define the purpose of its existence. 
They may arise from existing beliefs and contextual cues, such as detected user confusion or incomplete tasks, and can be assigned priorities, utilities, or activation conditions that influence their selection.
Last, \textit{Intentions} are desires to which the agent has committed and represent its active goals and responsibilities. 
For example, in an educational environment containing multiple agents, one agent may be assigned responsibility for assisting users with navigation. 
Intentions guide behavior over time and prevent frequent goal-switching in response to minor environmental changes.

\textbf{\textit{Reasoner Component}}:
This component is responsible for operating over the Cognitive Construct component. 
Specifically, this component retrieves the beliefs, desires, and intentions of the agents and generates dynamic structures that determine whether an agent is actually interested in a problem. 
Additionally, this component evaluates current beliefs to generate or update desires, assesses competing desires based on priorities or utilities, and selects intentions that the agent commits to pursuing. 
This process may involve the use of rule-based inference, logical reasoning, and constraint satisfaction algorithms. 
However, this is tightly dependent on the complexity and the nature of the application. 
Last, this component also manages intention revision, allowing the agent to drop or replace intentions when beliefs change significantly or when goals are achieved. 

\textbf{\textit{Planner Component}}:
This component is responsible for transforming the agent’s beliefs, desires, and intentions into structured sequences of actions, as a means to enable the agent to achieve its goals. 
It operates at an abstract level, reasoning about symbolic actions, their preconditions, and their expected effects. 
Depending on the complexity of the domain, the Planner may employ goal-oriented action strategies and planning. 
The output of the Planner is a plan consisting of ordered or partially ordered actions that represent what the agent should do. 

\textbf{\textit{Act Component}}:
This component retrieves from the Think component the decision on what the agent should do. 
Subsequently, it makes this decision and communicates with the Behavior Mapper component to plan how to perform an action based on the agent’s capabilities. 
Once the how is determined, then this component performs that action in the environment. 
Effectively, this component serves as the final interface between the agent’s cognition and the external world (i.e., the environment), by triggering animations, speech outputs, navigation, etc. 
The Act component operates in real time and must respect performance and synchronization constraints imposed by the environment. 
As a result, the operations need to be optimized toward complexity and performance. 
For that reason, to operate this component, different forms of finite state machine algorithms can be used.

\textbf{\textit{Behavior Mapper Component}}:
This component is responsible for retrieving the actions needed from the agent to be performed, and at the same time, it requests from the Agent Capabilities relevant capabilities, so the final behavior of the agent can be planned. 
Effectively, this component is responsible for transforming abstract planned actions into concrete executable behaviors. 
To perform this, this component translates symbolic action descriptions produced by the Think component into specific behavior specifications that are compatible with the agent’s capabilities. 
This mapping process may involve selecting appropriate animations, generating dialogue acts, and choosing gesture variants, among others.  
As a result, this component ensures that the agent’s internal intentions are expressed in a believable embodied form.

\textbf{\textit{Agent Capabilities Component}}:
This component is responsible for defining the set of actions that an agent can realistically perform within the virtual environment. 
These capabilities reflect the agent’s embodiment. 
As a result, this component behaves as a central database that includes all the animations and behaviors of the agent, including body movements, locomotion mechanisms, facial expressions, voice recordings, gaze, and gestures, among many others. 
This component is in constant communication with the Behavior Mapper component. 
This is because, once an action is decided, the mapper should map the necessary capabilities of the agent, and for that reason, this component provides the mapper component with the relevant resources, with the use of search algorithms.

\section{Prototype-Based Feasibility Evidence}
CEAA informed the implementation of pedagogical-agent-driven learning environments (two identical environments: one for computer desktops and one for VR) developed in Unity3D, representing a virtual museum dedicated to ENIAC and early computing history.
The museum contained learning materials, an interactive three-dimensional ENIAC representation, and four embodied pedagogical agents assigned with different pedagogical responsibilities. 

The agents served as navigators and instructors, delivering presentations on ENIAC’s historical development, components, operation, and limitations. 
They also supported the learning process by behaving as evaluators, answering learners’ questions and conducting Q\&A-based assessment activities that provided adaptive feedback. 
To implement these agents, CEAA was used to enable them to reason about their goals, select contextually appropriate actions, and perform these actions in the learning environment. 
Their embodied capabilities included speech, gaze, gestures, facial expressions, body movement, spatial awareness, navigation, obstacle avoidance, and continuous monitoring of learner progress and the state of the environment. 
The agents also communicated with one another to coordinate their roles and activities within the shared virtual environment.

From a CEAA perspective, the prototype facilitated the interaction between the environment, shared knowledge, and autonomous agent processes. 
Learner actions, object interactions, presentation progress, assessment responses, and system events originated in the environment and were reflected in the maintained state of the learning experience. 
Contextual and episodic information enabled the agents to relate current events to earlier interactions. 
Agent-specific goals and roles then guided reasoning, planning, behavior selection, and embodied action. 
For example, completing an instructional presentation could update the learner’s progress, enable the teaching-assistant agent to provide follow-up support, and subsequently allow the evaluator agent to initiate an assessment. 
Similarly, episodic memory enabled an agent to recognize that a presentation had already been delivered and adapt its subsequent response. 

The implementation was evaluated through a mixed-methods comparative study involving 92 undergraduate Computing and Engineering students, evaluating the differences in user learning outcomes and their acceptance of the technology to support their learning experience. 
Participants were randomly allocated to either the 3D desktop or immersive VR condition, with 46 participants in each group. 
Each participant completed a knowledge pre-test, interacted with the assigned environment for approximately 45 minutes, completed a post-test, and responded to a post-experience questionnaire measuring effort expectancy, performance expectancy, behavioral intention to use, and attitude towards use. 
In addition, 12 participants from the VR condition participated in structured focus-group discussions. 
The evaluation results provide evidence VR and desktop implementations of the pedagogical agents supported statistically significant knowledge acquisition. 
All participants demonstrated positive pre-test-to-post-test gains in learning outcomes, confirming significant within-group improvements. 
Although the aggregate learning gain was slightly higher in the desktop condition than in VR, the between-group difference was not statistically significant, indicating that the same agent-supported educational design could facilitate learning effectively across both delivery modes.

Technology acceptance was also positive in both conditions. 
Participants rated both systems as relatively easy to use and useful for learning and reported positive attitudes and intentions to use them in future educational activities. 
The VR condition produced slightly higher median values for effort expectancy and behavioral intention, and both conditions obtained similar median values for performance expectancy and attitude towards use. 
The focus-group findings complemented these results by indicating that participants associated the VR experience with interactivity, immersion, engagement, and information retention.
These findings demonstrate that a multi-agent, knowledge-aware and embodied learning system could be implemented and used successfully in a realistic educational activity.
Accordingly, the study demonstrates that CEAA’s underlying principles can support the construction of deployable multi-agent VR and desktop experiences, but further technical evaluation is required to establish the architecture’s performance, scalability, and generalizability across application domains.

\section{Discussion}
The architecture presented in this work responds to the practical challenges of deploying agents with cognitive capabilities in real-time virtual environments. 
Its feasibility has been demonstrated through prototype learning environments populated by pedagogical agents with different educational roles \cite{hadjiliasi2024comparative}. 
Although several existing architectures provide strong theoretical foundations, many remain highly abstract and difficult to integrate into real-time 3D environments because of their complexity and dependence on specific engines or interaction paradigms. 
The proposed architecture addresses this gap by providing an implementation-oriented framework that enables agents to perform cognitive functions, make run-time decisions, and simulate human-like intelligence and behavior. 
It integrates multiple architectural approaches into an easy-to-follow, scalable, and maintainable solution for developing embodied IVAs. 
Building on the sense-think-act paradigm, explicit memory processing, blackboard-based knowledge management and representation, pandemonium-inspired algorithms, and BDI reasoning, it provides a modular “brain” template for IVAs operating in complex, dynamic, real-time environments, including VR, Metaverse applications, serious games, and CPSS.

\subsection{Development and Implementation Considerations}
Although CEAA is implementation-oriented, its deployment in real-time interactive systems requires several practical considerations. 
First, latency should be managed by separating frame-critical processes from computationally expensive cognitive operations. 
Environment sensing, behaviour triggering, animation control, and action execution should remain lightweight and close to the environment’s real-time update loop, whereas memory retrieval, reasoning, and planning may run asynchronously or through event-driven updates, such as when an event relevant to the agent occurs. 
This prevents the cognitive cycle from blocking rendering, physics, or user interaction.
Second, memory overhead should be controlled by distinguishing between short-term event logs, agent-specific memories, and long-term knowledge. 
The shared knowledge base should store only information relevant to the agents inhabiting the environment rather than every environmental event, thereby reducing unnecessary computation and memory use.
Third, multi-agent scalability requires avoiding duplicated processing. 
The shared knowledge base can reduce redundancy by maintaining a common symbolic representation of the environment, while each agent selectively processes events relevant to its beliefs, desires, intentions, role, or capabilities.
Integration with engines such as Unity and Unreal can be achieved by mapping CEAA components to engine-level structures. 
The Environment and Act components can interface with scene objects, physics, animation, and interaction systems, while the Knowledge Base, Memory, Reasoner, and Planner can operate as independent services or manager components. 
The Behaviour Mapper can then translate abstract decisions into engine-specific animations, dialogue, navigation commands, or other interaction scripts. 
CEAA should therefore be understood not as a computationally fixed implementation, but as a deployment template whose real-time feasibility depends on asynchronous execution, selective perception, bounded memory, and modular integration adapted to the target engine and environment.

\subsection{Contributions}
The primary contribution of CEAA is an implementation-oriented architectural framework for deploying IVAs that exhibit cognitive, goal-directed, adaptive, and embodied behaviour in real-time virtual environments. 
It addresses the persistent gap between cognitive agent models and their practical deployment by translating established theoretical concepts into modular components that can be implemented within interactive 3D systems. 
Unlike existing architectures that often examine reasoning at a conceptual level, CEAA integrates multiple approaches into a development-oriented framework that explicitly maps cognitive processes to deployable IVA components. 
Its contribution therefore lies not in proposing a new reasoning model, but in fusing established frameworks into a coherent implementation structure.
Instead of providing a high-level abstraction, CEAA decomposes cognition and decision-making into concrete, interconnected modules that can be mapped to complex real-time systems, including VR applications, serious games, Metaverse applications, and CPSS. 
This level of specification reduces implementation uncertainty, supports modular development, testing, scalability, and explainability, and facilitates deployment in modern game and simulation engines such as Unity and Unreal Engine.

CEAA also contributes by integrating cognition and embodiment within a single decision-making process in which both have equal importance. 
Whereas many architectures primarily emphasize cognition and treat embodiment as an output, CEAA embeds behavior expression within the decision-making pipeline. 
This is particularly relevant to IVA research, where the generation of expressive and contextually appropriate behavior remains challenging.
Furthermore, CEAA provides a reusable and extensible reference model rather than a domain-specific solution. 
It functions both as a theoretical framework and as a generalizable architectural backbone that can support IVA development across multiple application domains.
The novelty of CEAA is therefore architectural and implementation-oriented. 
It does not replace established approaches such as BDI, blackboard architectures, Sense–Think–Act, or common game-AI structures, but organizes them into a sequential pipeline for embodied IVAs operating in real-time interactive 3D environments. 
CEAA extends the generic Sense–Think–Act loop by separating sensing, memory, reasoning, planning, behavior mapping, and action execution into distinct but connected components.
It differs from standalone BDI architectures by linking beliefs, desires, and intentions with shared environmental knowledge, agent-specific memory, capability-aware planning, and embodied behavior execution. 
Although it adopts the blackboard principle for shared knowledge representation, the blackboard forms only one layer of the broader architecture rather than the complete control mechanism. 
CEAA’s contribution consequently lies in translating and aligning established cognitive and game-AI concepts into a reusable structure that supports practical deployment, modularity, explainability, and embodiment-aware agent behavior.

\section{Conclusions, Limitations and Future Work}
This paper proposes CEAA, a modular, implementation-oriented architecture for deploying embodied IVAs with cognitive function to operate in real-time virtual environments.
This architecture was developed as part of a broader research project, and it aims to address the disconnection between the deployment of agents that have cognitive functions and their practical development in real-time interactive 3D systems. 
This architecture builds on several architectures and paradigms, and offers a modular framework that clearly separates the environmental dynamics, cognitive processes, and embodied execution for IVAs. 
This separation supports adaptive, goal-oriented, and explainable agent behavior without sacrificing real-time performance. 
The architecture’s modular design aligns with both object-oriented and component-based development paradigms, and as a subsequent event, it enables a straightforward implementation solution in modern game engines such as Unity and Unreal. 
As a result, it offers a template and a backbone of a reusable and extendable framework that minimizes the barrier for deploying IVAs with cognitive capabilities in complex interactive computing environments.

Despite the advantages of the proposed architecture, like any other system, it comes with some limitations. 
The most significant limitation is that although the architecture has been used to test its feasibility in prior studies \cite{hadjiliasi2024comparative}, it remains at a conceptual and theoretical level. 
As such, further studies are planned to benchmark this architecture and compare its capabilities with other architectures, such as its components (e.g., BDI, sense-think-act, etc.), across multiple dimensions, including latency, ease of development, user experience, etc.
Additionally, another limitation is that this architecture, at its current stage, is still at a conceptual and abstract level. 
Hence, it can only be used as a point of reference and guidelines for others to implement the “brains” for their agents. 
However, it could benefit other developers and researchers if a tool that implements this architecture exists. 
For that reason, future work focuses on the implementation of extension tools for Unity and Unreal that implement this architecture and effectively ease the development of agents.



\bibliographystyle{ieeetr}
\bibliography{COMPSAC2026}    
    

\end{document}